\documentclass[runningheads]{llncs}

\usepackage{eccv}

\usepackage{eccvabbrv}

\usepackage{graphicx}
\usepackage{booktabs}

\usepackage[accsupp]{axessibility}  

\usepackage{float}
\usepackage{booktabs}
\usepackage{multirow}
\usepackage{bbding}
\usepackage{makecell}
\usepackage{cancel}
\usepackage{color}
\usepackage{wrapfig}
\usepackage{enumitem}

\usepackage{hyperref}

\usepackage{orcidlink}
\newcommand{\minisection}[1]{ \noindent {\bf #1}\ \ }

\begin{document}

\title{Class-Incremental Learning with CLIP: Adaptive Representation Adjustment and Parameter Fusion} 

\titlerunning{CIL with CLIP: Adaptive Representation Adjustment and Parameter Fusion}

\author{Linlan Huang\inst{1}\orcidlink{0009-0007-0515-1202} \and
Xusheng Cao\inst{1} \and
Haori Lu\inst{1}\orcidlink{0000-0003-0167-6646} \and
Xialei Liu\inst{1,2}\orcidlink{0000-0001-8534-3026}\textsuperscript{(\Envelope)}}

\authorrunning{L.~Huang et al.}

\institute{$^1$VCIP, CS, Nankai University \\
$^2$ NKIARI, Shenzhen Futian \\
\email{\{huanglinlan, caoxusheng, luhaori\}@mail.nankai.edu.cn} \\
\email{\{xialei\}@nankai.edu.cn} \\ 
}

\maketitle

\begin{abstract}
Class-incremental learning is a challenging problem, where the goal is to train a model that can classify data from an increasing number of classes over time.
With the advancement of vision-language pre-trained models such as CLIP, they demonstrate good generalization ability that allows them to excel in class-incremental learning with completely frozen parameters. However, further adaptation to downstream tasks by simply fine-tuning the model leads to severe forgetting. 
Most existing works with pre-trained models assume that the forgetting of old classes is uniform when the model acquires new knowledge.
In this paper, we propose a method named Adaptive \textbf{R}epresentation \textbf{A}djustment and \textbf{P}arameter \textbf{F}usion (\textbf{RAPF}). During training for new data, we measure the influence of new classes on old ones and adjust the representations, using textual features.
After training, we employ a decomposed parameter fusion to further mitigate forgetting during adapter module fine-tuning.
Experiments on several conventional benchmarks show that our method achieves state-of-the-art results.
Our code is available at \url{https://github.com/linlany/RAPF}.
\end{abstract}

\section{Introduction}
\label{sec:intro}

The real world is constantly changing, which requires the model to adapt to new knowledge while retaining old knowledge. 
If not updated, the models can become obsolete, degrading their performance over time~\cite{geng2020recent}. 
Privacy and storage constraints may limit access to old data, leading to a severe imbalance in data distribution. 
This imbalance, when models are updated, causes models to become biased towards current data and forget previously acquired knowledge, a phenomenon known as catastrophic forgetting~\cite{li2017learning}.
Therefore, the challenge of continuous learning is to balance plasticity and stability~\cite{masana2022class}, allowing models to learn new knowledge without forgetting old knowledge and to reuse and expand knowledge from experience across different tasks.

\begin{figure}
    \centering
    \includegraphics[width=0.8\linewidth]{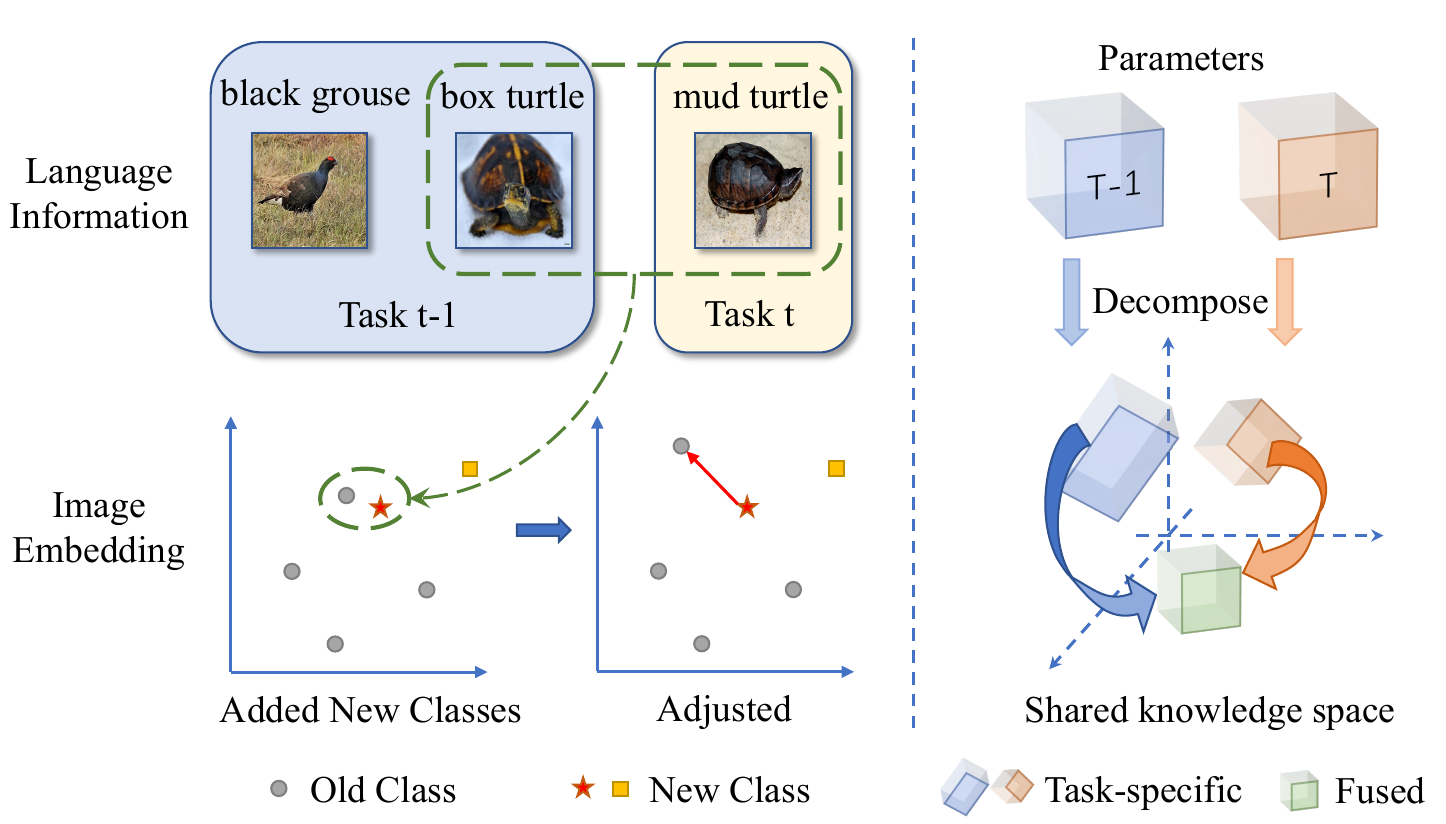}
    \caption{{Semantically similar categories pose significant challenges in CIL across tasks, in which language information can help to pick out the adjacent old and new classes when new data are encountered. Then the image feature representation of the old category can be adjusted accordingly. 
    Additionally, a decomposed parameter fusion strategy is further adapted to reduce forgetting. 
    We decompose the parameters learned from two consecutive tasks into shared knowledge and task-specific knowledge. Then, we fuse the parameters based on this decomposition.}}
    \label{fig:intro}
\end{figure}

Class-incremental learning, a challenging aspect of continual learning, has garnered increasing attention recently.
It involves learning from a data stream where new classes are added over time.
To cope with this scenario, three main categories of methods have been proposed: regularization-based, replay-based, and parameter isolation-based methods~\cite{de2021continual}. 
These methods aim to preserve previous knowledge by adding regularization terms to the loss function, replaying old data samples, or allocating dedicated parameters for each class. 
However, most of these methods rely on models that are trained from scratch, which may not be optimal for incremental learning. 
Thus, an important research direction is to reduce forgetting by applying various methods to models pre-trained on large-scale datasets.

Pre-trained models on large-scale datasets have shown remarkable generalization ability and robustness against catastrophic forgetting in downstream tasks~\cite{lee2023pre,wu2022classs_base_trong,mehta2021pretrain_invest,wang2022s-prompt}. 
Moreover, the visual language pre-trained model (CLIP)~\cite{radford2021clip} has demonstrated powerful zero-shot ability in continual learning~\cite{thengane2022continualclip} and high adaptability to downstream tasks~\cite{gao2023clip-adapter, zhou2022coop, wang2024unlockingmultimodalpotentialclip}. 
Leveraging the excellent feature extraction ability of the pre-trained model, each incremental step requires updating only a small number of parameters, reducing the risk of forgetting. In contrast, models trained from scratch lack this advantage and may suffer from severe performance degradation. 
Therefore, pre-trained models have achieved impressive results in continual learning. 
There are two main strategies for continual learning with pre-trained models: fine-tuning the model~\cite{zhang2023slca} or continuously expand a small number of parameters on the model, such as methods based on prompts~\cite{smith2023coda,wang2022dualprompt,wang2022l2p,wang2022s-prompt} or adding adapters~\cite{zhou2023adam}.

Fine-tuning models may compromise the feature extraction ability of the original model and cause catastrophic forgetting, even with regularization constraints. 
Expanding parameters may mitigate interference with the original model but increase time and space costs over time. 
For the visual language pre-trained model, the language encoder provides rich information beneficial for continual learning.
However, most existing methods only use text features for classification, not fully exploring their potential to reduce forgetting.

In this paper, we propose a method that uses text features to enhance the classification ability of neighboring categories in class-incremental learning. 
It is based on the observation that CLIP relies on the fixed text features for classification.
The text features determine the decision boundary. 
When a new category arrives, the new decision boundary may divide part of the old category sample into the new category.
To address it, we need to enhance the separation of neighboring categories.
Using the text features, we can infer the relationship between the old and new categories.  
We select the pairs of neighboring categories by calculating the distance between the textual features of the new and old categories. 
Since the new category has sufficient data for learning, we do not need to modify its representation. Instead, we focus on adjusting the representation of old categories that are affected by new categories, as illustrated in \cref{fig:intro}.
We train a single linear layer to prevent compromising the feature extraction capability of the pre-trained model.

Moreover, we propose a decomposed parameter fusion method for the linear adaptive layer that does not increase the number of parameters as the task increases. 
Unlike directly calculating parameter averages, our fusion strategy is more fine-grained and considers the shared knowledge between tasks.
To balance stability and plasticity, we merge the parameters before and after an incremental task according to the parameter changes caused by learning the current task. 
We do not need to add extra distillation loss in the training process to constrain the parameter changes, which lowers the training cost.

The main contributions of this paper are as follows:
\begin{itemize}
    \item We explore a method to reduce the forgetting of CLIP models by using the textual features of category names;
    \item We propose a simple but effective method of decomposed parameter fusion for the linear layer adapter of pre-trained models;
    \item We achieved state-of-the-art results on several datasets.
\end{itemize}

\section{Related Work}
\label{sec:relate}
\subsection{Class-Incremental Learning (CIL)}
Incremental learning methods can be classified into three types according to the main strategy they use~\cite{de2021continual}. 
The methods based on regularization suppress the forgetting of old categories by imposing some constraints on the model’s parameters or outputs. MAS~\cite{aljundi2018MAS} and EWC~\cite{kirkpatrick2017ewc} compute the importance of each parameter for the old task and then add a regularization term to avoid catastrophic forgetting. Coil~\cite{zhou2021coil} uses knowledge distillation based on optimal transport to share knowledge between tasks. The methods~\cite{douillard2020podnet,li2017lwf,zhang2020DMC} use the loss of knowledge distillation as the regularization term.
The methods based on replay preserve the memory of old categories by saving some samples or features of old categories (exemplar), and then replaying them with the new category samples when training new categories.
Some are exploring how to reduce the storage occupancy of exemplars~\cite{luo2023exemplarcompression}, while others are looking for better strategies to select samples that will be added to exemplars~\cite{liu2020mnemonics,lopez2017gem,sun2023Second_Order_select}. There are also methods that save additional models and samples to assist in the current model training~\cite{chaudhry2019er,rebuffi2017icarl, sarfrazerror2023error_sensi_replay}.
The methods based on dynamic architecture dynamically adjust the model’s structure to adapt to the learning of new categories~\cite{douillard2022dytox,wu2022classs_base_trong, wang2022foster,yan2021der}.

\minisection {CIL with Pre-trained Model.}
There are two main types of methods for using pre-trained models. One type is to fine-tune the parameters of the model itself to adjust the feature representation.
ZSCL~\cite{zheng2023ZSCL} utilizes numerous external data to distill the pre-trained model to maintain a stable feature space.
Zhang et al.~\cite{zhang2023slca} employ different learning rates to update the pre-trained backbone network and the classifier. 
The other type is to keep the pre-trained model unchanged and add parameters to adjust the feature representation. 
Liu et al.~\cite{liu2023class_clip_ada} introduce an adapter to the pre-trained CLIP model to adapt to incremental tasks. 
PROOF~\cite{zhou2023proof} trains an adapter for each task and uses cross-modal attention to fuse the language and vision information of CLIP.
RanPac~\cite{mcdonnell2024ranpac} uses a high-dimensional projection to separate features.
The recently proposed prompt-based methods ~\cite{smith2023coda,wang2022dualprompt,wang2022l2p,khan2023prompt_language} select the corresponding prompts to add to the model according to the output of the features by the pre-trained model and then re-obtain the features for classification. Among them, LGCL~\cite{khan2023prompt_language} tries to introduce language guidance.
Ostapenko et al.~\cite{ostapenko2022pretrain_replay} add a classification network to the pre-trained model, and use the pre-trained latent feature space for replay.

\minisection {CIL without Exemplar.}
Sometimes it is not possible to store old class samples due to privacy and memory constraints\cite{smith2023rehearsal_free}.
Some existing works without exemplars use Gaussian distribution to model data and help with classification~\cite{hayes2020Gaussian_linear,tang2023no_strong_prompt, zhang2023slca}. Other approaches include oversampling prototypes~\cite{zhu2022ssre} or augmenting prototypes to simulate replay samples~\cite{malepathirana2023napa_vq,zhu2021pass}.
Some recent works use models to synthesize data from the old task as a substitute for exemplars~\cite{choi2021dual_teacher_datafree,gao2022datafree}. Prompt-based methods use the frozen backbone and the relative isolation of prompt parameters to avoid using exemplars~\cite{smith2023coda,wang2022dualprompt,wang2022l2p,khan2023prompt_language}.

\begin{figure*}[ht]
  \centering
  \includegraphics[width=0.9\linewidth]{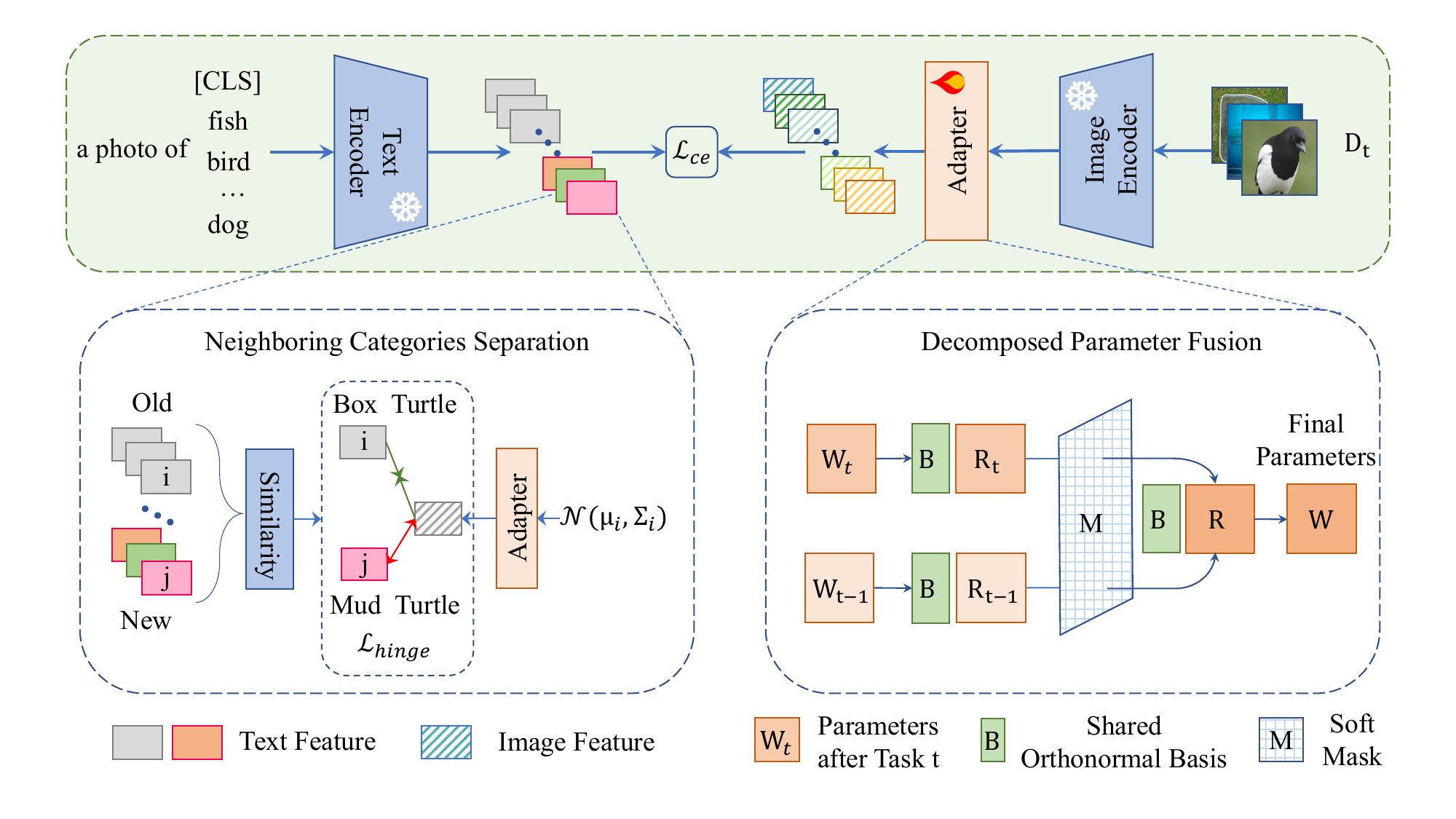}
  \caption{{The framework of our method. 
The neighboring categories separation module computes the similarity of text features to identify neighboring categories. We sample the distribution of the old class and calculate the hinge loss. 
In the parameter fusion module, we first decompose $\mathbf{W}_{t}$ and $\mathbf{W}_{t-1}$ into the same standard orthogonal basis $\mathbf{B}$. Then, we calculate a soft mask $\mathbf{M}$ from the difference of the decomposed parameters $\mathbf{R}_t$ and $\mathbf{R}_{t-1}$, which acts as the fusion weight. Finally, we reconstruct the parameter $\mathbf{W}$  from the fused parameter $\mathbf{R}$ and the basis $\mathbf{B}$.
  }}
  \label{fig:pipeline}
\end{figure*}

\section{Preliminaries}
\minisection {Class Incremental Learning Definition.}
A class-incremental learning algorithm trains a model $M_t$ at each task $t$, which can classify data from all classes that the algorithm has seen so far without task id, i.e., $C_1 \cup C_2 \cup ...\cup C_t$. And the category sets do not intersect, $\forall i \neq j, C_i\cap C_j = \emptyset$.
The model $M_t$ only uses the data in the current dataset $D_t$ to update from the previous model $M_{t-1}$, without accessing previous datasets.

\minisection {CLIP Adapter.}
An efficient method to adapt a pre-trained vision-language model to downstream tasks involves incorporating a small network as an adapter~\cite{gao2023clip-adapter}.
We denote the visual feature extractor of the CLIP backbone network as $f_{img}(\cdot)$, the text feature extractor as $f_{text}(\cdot)$, and the linear adapter as $A(\cdot)$. Given the image input $\mathbf{x}_i$, the class $y_i$ with the fixed prompt template, e.g., ``a photo of a [CLS]'', denoted by $\mathbf{t}_i$, the output result is as follows:
\begin{equation}\label{eq:output}
    p(y_i|\mathbf{x}_i) = \frac{\exp({\cos({A(f_{img}(\mathbf{x}_i)), f_{text}(\mathbf{t}_i)})/\boldsymbol{\tau}})}{\sum_{j=1}^{|y|} \exp(\cos(A(f_{img}(\mathbf{x}_i)),  f_{text}(\mathbf{t}_j))/\boldsymbol{\tau})},
\end{equation}
where $\boldsymbol \tau$ is the temperature. 
If the text encoder is frozen, we only need to preserve the embeddings of the text instead of class names in CIL.
We employ the cross-entropy loss criterion to fine-tune the parameters of the adapter, formulated as follows:
\begin{equation}\label{eq:loss_ce}
    \mathcal{L}_{ce}(\mathbf{y}, \mathbf{p}) = -\sum_{i=1}^n y_i \log {p}_i .
\end{equation}

\minisection {Feature Generation.}
Leveraging the well-learned representations by the pre-trained backbone, we can approximate the feature distribution of each category using a Gaussian distribution~\cite{zhang2023slca}.
We compute the centroid $\boldsymbol{\mu}_c$ and the covariance matrix $\boldsymbol{\Sigma}_c$ from the image embeddings denoted by $\mathbf{e}_c=f_{img}(\mathbf{x}_c)$ of class $c$. Then we can generate image embeddings for class $c$ by sampling from the Gaussian distribution $\mathcal{N}(\boldsymbol{\mu}_c, \boldsymbol{\Sigma}_c)$.

\section{Method}
\subsection{Overview}
As illustrated in \cref{fig:pipeline}, our framework comprises a pre-trained CLIP model and a linear adapter. 
The image features are derived from passing the images through the image encoder and adapter, while the label features are derived from passing the class labels through the text encoder. 
The classification result is determined by measuring the similarity between them. 
The old class features are generated by sampling from their respective Gaussian distributions and fed into the adapter along with the new data to compute classification loss.
Simultaneously, we select pairs of neighboring classes based on their text feature similarity. 
For each old class in these pairs, we sample more data from its Gaussian distribution and feed them into the adapter but only compute the loss for separation.
This way, we can adjust the feature representation of the old classes affected by the similar new classes, thereby reducing the forgetting of the old classes caused by learning the new classes.
Upon completion of the t-th task training phase, we fuse the parameters of the adapter from the preceding phase and the current adapter to obtain the ultimate adapter for this stage. The adaptive fusion of parameters can better balance plasticity and stability.
\subsection{Enhancing Neighboring Categories Separation with Language Guidance}

One manifestation of catastrophic forgetting is that the model erroneously identifies old category data as a new category. 
This phenomenon is more severe when the new category is similar to the old category.
These categories belong to neighboring categories.
They usually have similar semantic meanings in language and similar appearances, such as  `macaw' and `lorikeet'.
It is challenging for the model to distinguish them, especially when new categories are added.

By using the CLIP text encoder, we can measure the similarity between category names and use it to guide the adapter learning process. To select the nearest pairs of categories, we use the normalized text features of the category names to calculate the distance between the new and old categories:
\begin{equation}\label{eq:ydist}
    \mathbf{D} = \mathrm{dist}(f_{text}(\mathbf{t}_{new}),f_{text}(\mathbf{t}_{old})),
\end{equation}
where $\mathbf{D}$ is the matrix of Euclidean distances between the normalized text features of the new and old categories, $f_{text}(\mathbf{t}_{new})$ and $f_{text}(\mathbf{t}_{old})$. 
Learning new categories interferes with the performance of old ones. We focus on measuring the distance between the new and old categories, which reflects the degree of interference.
Then, we pick out the neighboring categories set $\mathcal{P}$:
\begin{equation}\label{eq:filter}
    \mathcal{P} = \{(i,j) | \mathbf{D}_{ij} < \boldsymbol{\alpha}\},
\end{equation}
where $\boldsymbol{\alpha}$ represents the threshold. $\mathbf{D}_{ij}$ is the distance between the text features of the old category $i$ and the new category $j$. 
This criterion reduces the complexity of many-to-many relationships by selecting a subset of one-to-one relationships.

We introduce a hinge loss by sampling the features of the old category from a normal distribution for each pair of neighboring categories.
Since we do not have real data on the old classes, excessive adjustments may harm the classification performance of the old classes. Therefore, we only use hinge loss to make small but effective adjustments.
The sampled data of the old category $c$ is indicated by $\hat{\mathbf{e}}_c$, and the adapter is indicated by ${A}$:
\begin{equation}\label{eq:l_hinge}
       \mathcal{L}_{hinge} = \sum_{k=1}^{|\mathcal{P}|} \max(\mathrm{dist}({A}(\hat{\mathbf{e}}_c),f_{text}(\mathbf{t}_c))- \mathrm{dist}( {A}(\hat{\mathbf{e}}_c),f_{text}(\mathbf{t}_{\cancel{c}}))+m,0), 
\end{equation}
where $m$ is a constant margin, $c$ and $\cancel{c}$ denote the old category and the new category in the $k$-th pair of neighbor categories which belong to set $\mathcal{P}$. 
This loss function can make the adapter focus more on the neighboring categories that are more likely to be confused while minimizing the disturbance to the representation of the old categories.

The final loss function is combined with Cross Entropy loss $\mathcal{L}_{ce}$ as:
\begin{equation}\label{eq:loss}
    \mathcal{L} = \mathcal{L}_{hinge} + \mathcal{L}_{ce}.
\end{equation}
\subsection{Increasing Stability with Decomposed Parameter Fusion}
We introduce a parameter fusion mechanism to maintain the stability of the linear adapter, thereby reducing forgetting. 

We evaluate the importance of each new parameter. The gradient indicates the direction of parameter updates based on new data, and the difference in the parameters between tasks is the weighted sum of the gradients. So the parameters with large changes are more important for learning new knowledge.

To obtain the importance matrix as a fine-grained weight for parameter fusion, we calculate the difference between the parameters after training and the parameters from the previous task and normalize it by the maximum value:
\begin{equation}\label{eq:mask}
    \mathbf{M} = \min(1, \frac{\lvert \mathbf{W}_{new} - \mathbf{W}_{old}\rvert}{\max(\lvert \mathbf{W}_{new} - \mathbf{W}_{old}\rvert)}+ b),
\end{equation}
where $\mathbf{M}$ denotes the importance of each new parameter, $\mathbf{W}_{new}$ denotes the parameters obtained from the current task training, $\mathbf{W}_{old}$ denotes the parameters from the previous task and $b$ is constant bias.
In order to compare the difference between $\mathbf{W}_{new}$ and $\mathbf{W}_{old}$ under the same standard, we decompose $\mathbf{W}_{old}$ to an orthonormal basis $\mathbf{B}$ by SVD~\cite{golub1971SVD} and calculate the projection from matrix $\mathbf{W}_{new}$ to $\mathbf{B}$:
\begin{equation}\label{eq:mdecompose}
    \mathbf{W}_{old} \xrightarrow{decompose} \mathbf{B} \mathbf{R}_{old},
\end{equation}
\begin{equation}\label{eq:mproj}
    \mathbf{R}_{new} = \mathbf{B}^{\mathsf{T}} \mathbf{W}_{new}.
\end{equation}
The parameter matrix $\mathbf{W}_{old}$ and $\mathbf{W}_{new}$ is expressed as linear combinations of the orthonormal matrix $\mathbf{B}$, i.e., $\mathbf{R}_{old}$ and $\mathbf{R}_{new}$, which represent task-specific knowledge.
The matrix $\mathbf{B}$ represents the shared knowledge space across the parameter matrices. Thus, the matrix difference that we calculate can be transformed into calculating the difference of two distinct weights of the same orthonormal basis.

We substitute the $\mathbf{R}_{new}$ and $\mathbf{R}_{old}$ obtained from the decomposition into the $\mathbf{W}_{new}$ and $\mathbf{W}_{old}$ of \cref{eq:mask} to calculate $\mathbf{\mathbf{M}}$.
Then we compute the matrix $\mathbf{R}$ and the final parameter $\mathbf{W}$:
\begin{equation}\label{eq:p_w_ema}
    \mathbf{R} = (\mathbf{J}-\mathbf{M}) \odot \mathbf{R}_{old} + \mathbf{M} \odot \mathbf{R}_{new},
\end{equation}
\begin{equation}\label{eq:mmul}
    \mathbf{W} = \mathbf{B} \mathbf{R},
\end{equation}
where $\mathbf{J}$ denotes a matrix of all ones and $\odot$ indicates the element-wise product.

\section{Experiments}
\subsection{Experimental Setup}
\minisection {Datasets.}
We conduct our experiments using three datasets: CIFAR00~\cite{krizhevsky2009cifar100}, ImageNet1K~\cite{deng2009imagenet}, ImageNet100, ImageNet-R~\cite{hendrycks2021imagenet-r} and CUB200~\cite{cub200}.
The CIFAR100 dataset consists of 100 categories. Each category contains 600 color images with a resolution of 32 × 32 pixels. Of these, 500 images are assigned to the train set, and 100 images are assigned to the test set.
The ImageNet1K dataset consists of 1000 categories and the ImageNet100 dataset is a subset of the ImageNet1K dataset that consists of 100 selected classes. 
The ImageNet-R dataset is a collection of diverse image categories derived from the ImageNet dataset. The ImageNet-R dataset includes images in various styles such as art, cartoons, graffiti, embroidery, video games, and so on. They are different expressions of the 200 categories in the ImageNet dataset. We follow the work ~\cite{thengane2022continualclip,wang2022dualprompt} to split the dataset into a training set and a test set.
The CUB200~\cite{cub200} dataset is widely used for fine-grained visual categorization tasks. It comprises 11,788 images of 200 subcategories belonging to various species of birds.

\minisection {Competing methods.}
We compare to CIL methods: L2P ++~\cite{wang2022l2p}, DualPrompt~\cite{wang2022dualprompt}, CODA~\cite{smith2023coda}, SLCA~\cite{zhang2023slca}, ADAM-Adapter~\cite{zhou2023adam} and PROOF~\cite{zhou2023proof}. 
Continual-CLIP~\cite{thengane2022continualclip} refers to the zero-shot performance of the CLIP model. 
PROOF~\cite{zhou2023proof} is the method that is based on CLIP with exemplar.
To ensure fairness of comparison, all methods use the same OpenAI CLIP pre-trained weights~\cite{radford2021clip}.
The results of the three methods DualPrompt, L2P ++, and CODA are obtained by running the publicly available code implementation of method CODA.
The results of SLCA, ADAM-Adapter, PROOF and Continual-CLIP methods are obtained from their respective public code.

\begin{table*}[t]
    \centering
    \caption{{Experimental results for continual learning on CIFAR100. B denotes the number of base classes, and Inc denotes the number of incremental classes. All baseline-based results are reproduced according to their published code with CLIP pre-trained weights for VIT-B/16. We run our experiments for several different shuffles of the class order and report the mean of these orders.}}
    \label{tab:cifar100_result}
\resizebox{0.99\textwidth}{!}{
    \begin{tabular}{cccccccccccc}
    \toprule%
    \multirow{2}{*}{Method}
    & \multirow{2}{*}{Exemplar}
    &\multicolumn{2}{c}{B0 Inc5} 
    &\multicolumn{2}{c}{B0 Inc10} 
    &\multicolumn{2}{c}{B0 Inc20} 
    &\multicolumn{2}{c}{B50 Inc5} 
    &\multicolumn{2}{c}{B50 Inc10}\\
    &&Avg &Last &Avg &Last &Avg &Last &Avg &Last &Avg &Last\\
    \midrule
    PROOF~\cite{zhou2023proof}           &\Checkmark    &\underline{85.12}&\underline{76.13} &\underline{84.88}&\underline{76.29}        &84.11&76.86 &83.22&76.25 &83.17&76.5\\
    \midrule%
    L2P ++~\cite{wang2022l2p}                        
    &\XSolidBrush 
    &79.18&68.67  &81.90&73.08  &84.39&77.37 &58.57&18.04     &76.51&48.52\\
    DualPrompt~\cite{wang2022dualprompt}             
    &\XSolidBrush 
    &79.74&69.91  &81.45&72.51  &85.19&\underline{77.47} &58.55&15.26     &72.00&45.05 \\
    CODA~\cite{smith2023coda}                        
    &\XSolidBrush 
    &69.78&41.98  &76.98&62.25  &78.65&65.29 &58.45&15.99     &67.88&28.77\\
    Continual-CLIP~\cite{thengane2022continualclip}  
    &\XSolidBrush 
    &75.93&66.68  &75.15&66.68  &74.01&66.68 &70.79&66.68     &70.77&66.68 \\
    SLCA~\cite{zhang2023slca}                        
    &\XSolidBrush 
    &78.96&66.84  &80.53&67.58  &\underline{85.25}&76.99  &\textbf{86.99}&76.8     &\textbf{86.55}&\textbf{79.92}\\
    ADAM-Adapter~\cite{zhou2023adam}                 
    &\XSolidBrush 
    &70.18&58.12  &75.76&65.50  &77.28&67.89 &83.38&\underline{76.94}     &83.21&76.94\\
    \midrule%
    ours 
    & \XSolidBrush 
    &\textbf{86.87}&\textbf{79.26} &\textbf{86.19}&\textbf{79.04} &\textbf{85.73}&\textbf{79.24} &\underline{85.03}&\textbf{79.64} &\underline{84.73}&\underline{79.36}\\
    \bottomrule%
    
    \end{tabular}
}
\end{table*}

\begin{table*}[t]
    \caption{{Experimental results for continual learning on ImageNet100.}}
    \centering

    \label{tab:imagenet100_result}
\resizebox{0.99\textwidth}{!}{
    \begin{tabular}{cccccccccccc}
    \toprule%
    \multirow{2}{*}{Method}
    & \multirow{2}{*}{Exemplar}
    &\multicolumn{2}{c}{B0 Inc5} 
    &\multicolumn{2}{c}{B0 Inc10} 
    &\multicolumn{2}{c}{B0 Inc20} 
    &\multicolumn{2}{c}{B50 Inc5} 
    &\multicolumn{2}{c}{B50 Inc10}\\
    &&Avg &Last &Avg &Last &Avg &Last &Avg &Last &Avg &Last\\
    \midrule
    PROOF~\cite{zhou2023proof}           &\Checkmark     &\underline{86.92}&75.52   &84.71&72.48     &81.92&68.56
 &84.16&74.44 &82.78&71.04\\
    \midrule%
    L2P ++~\cite{wang2022l2p}           &\XSolidBrush              &75.43&62.10  &80.51&67.22  &84.12&73.70  &62.00&22.15    &74.11&49.46\\
    DualPrompt~\cite{wang2022dualprompt} &\XSolidBrush             &75.40&61.10  &80.65&67.38  &84.65&74.24  &62.10&22.36    &74.20&49.78 \\
    CODA~\cite{smith2023coda}           &\XSolidBrush              &51.64&24.94  &64.13&34.76  &69.78&43.96  &57.33&19.95    &65.14&28.80\\
    Continual-CLIP~\cite{thengane2022continualclip}  &\XSolidBrush &85.74&75.40  &84.98&75.40  &84.03&75.40  &81.35&75.40    &81.09&75.40 \\
    SLCA~\cite{zhang2023slca}            &\XSolidBrush             &78.40&63.36  &78.63&59.92  &84.08&71.08  &\underline{86.47}&72.22     &\underline{86.26}&71.18\\
    ADAM-Adapter~\cite{zhou2023adam}    &\XSolidBrush              &85.78&\underline{75.72}  &\underline{85.84}&\underline{76.40}  &\underline{85.85}&\underline{77.08}  &84.90&\underline{78.58}    &84.60&\underline{78.58}\\
    \midrule%
    ours &\XSolidBrush &\textbf{87.59}&\textbf{79.87} &\textbf{87.51}&\textbf{80.23} &\textbf{86.72}&\textbf{80.10} &\textbf{86.53}&\textbf{80.16} &\textbf{86.36}&\textbf{80.22}\\
    \bottomrule%
    
    \end{tabular}
}
\end{table*}

\minisection {Evaluation metrics.}
The average precision after training the t-th task in the test data for the first to t-th tasks is denoted as $A_t$. 
$Avg$ is the average of the accuracies of all tasks. 
$Last$ is the average accuracy after the last tasks.

\minisection {Implementation detail.}
We developed our method using PyTorch and ran it on an RTX 3090 GPU. Our backbone is the ViT-B/16 version of CLIP. We train the model with the Adam optimizer for 15 epochs, starting with a learning rate 0.001. We use a MultiStepLR scheduler that reduced the learning rate by a factor of 0.1 at epochs 4 and 10. 
We use 0.65 as the default threshold for the text feature distance to select the adjacent category pairs.
We used approximately 2000 sampled data per epoch to simulate replay samples, which matched the amount of data added by the conventional setting with replay. 
In each iteration, we sampled 20 additional features for every category that was selected by threshold.
Due to insufficient data for some classes, we cannot obtain a full-rank covariance matrix. We follow previous works~\cite{van1980shrinka,kumar2022shrinkb} and use covariance shrinkage to obtain a full-rank matrix.
We run our experiments for several different shuffles of the class order and report the mean of these orders. 

\begin{table*}[t]
    \caption{{Experimental results for continual learning on ImageNet-R.}}
    \centering
    \label{tab:imagenet-r_result}
\resizebox{0.99\textwidth}{!}{
    \begin{tabular}{cccccccccccc}
    \toprule%
    \multirow{2}{*}{Method}
    & \multirow{2}{*}{Exemplar}
    &\multicolumn{2}{c}{B0 Inc10} 
    &\multicolumn{2}{c}{B0 Inc20} 
    &\multicolumn{2}{c}{B0 Inc40} 
    &\multicolumn{2}{c}{B100 Inc10} 
    &\multicolumn{2}{c}{B100 Inc20}\\
    &&Avg &Last &Avg &Last &Avg &Last &Avg &Last &Avg &Last\\
    \midrule
    PROOF~\cite{zhou2023proof}           &\Checkmark   &\underline{82.69}&\underline{77.25}  &\underline{82.83}&\underline{77.05}        &82.63&77.12 &81.61&77.10 &81.78&77.17\\
    \midrule%
    L2P ++~\cite{wang2022l2p}           &\XSolidBrush             &76.87&68.78  &81.67&75.98  &82.81&77.87  &56.17&17.90    &67.73&43.28\\
    DualPrompt~\cite{wang2022dualprompt}  &\XSolidBrush           &77.07&69.41  &82.01&75.77  &\underline{83.77}&78.64  &57.37&19.18    &69.18&45.37 \\
    CODA~\cite{smith2023coda}           &\XSolidBrush             &75.23&64.53  &78.00&67.52  &78.80&71.27  &56.62&17.64    &65.62&35.06\\
    Continual-CLIP~\cite{thengane2022continualclip}&\XSolidBrush  &79.84&72.00  &79.12&72.00  &77.59&72.00  &76.93&72.00    &76.76&72.00 \\
    SLCA~\cite{zhang2023slca}            &\XSolidBrush            &80.18&73.57  &75.92&70.37  &83.35&\underline{79.1}   &\underline{82.85}&\underline{78.57}    &\underline{83.50}&\underline{79.67}\\
    ADAM-Adapter~\cite{zhou2023adam}    &\XSolidBrush             &76.71&68.75  &78.65&71.35  &79.87&73.02  &79.87&75.37    &79.75&75.37\\
    \midrule%
    ours&\XSolidBrush &\textbf{86.28}&\textbf{79.62} &\textbf{85.58}&\textbf{80.28} &\textbf{84.69}&\textbf{80.18} &\textbf{84.12}&\textbf{80.04} &\textbf{83.99}&\textbf{80.35}\\
    \bottomrule
    
    \end{tabular}
}
\end{table*}
\begin{figure*}[t]
    \centering
    \includegraphics[width=\linewidth]{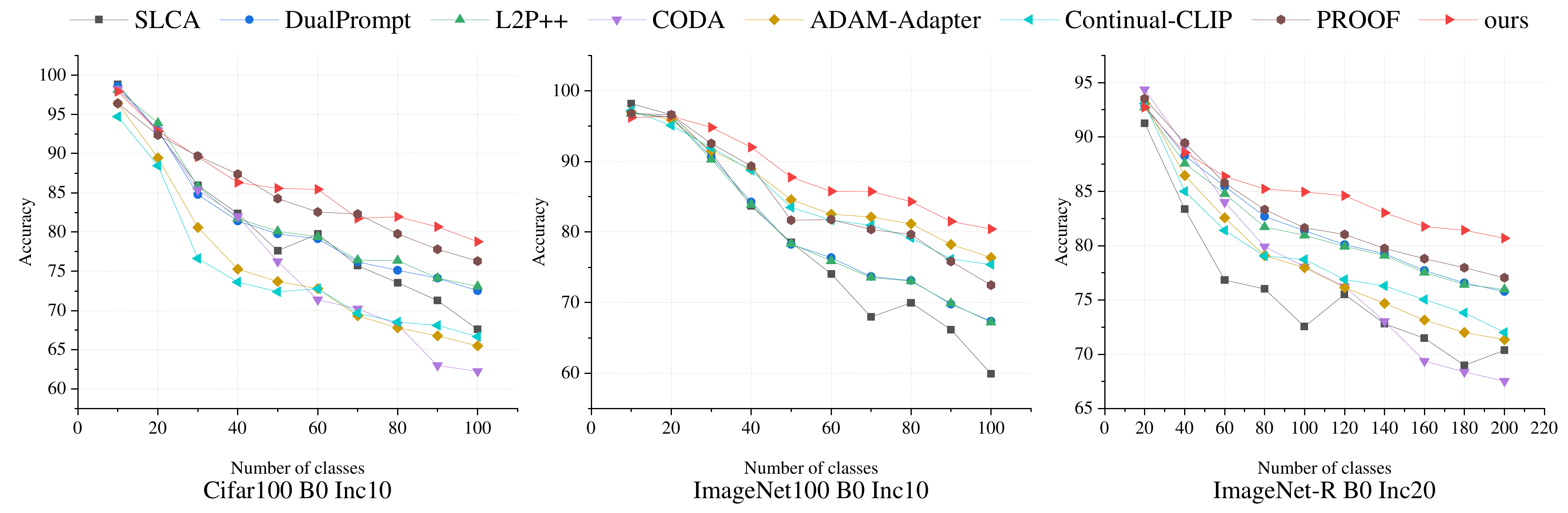}
    \caption{Accuracy curve of our method with other SOTA baselines on CIFAR100, ImageNet100 and ImageNet-R.}
    \label{fig:curve}
\end{figure*}

\subsection{Comparison Results}
\Cref{tab:cifar100_result}, \cref{tab:imagenet100_result} and \cref{tab:imagenet-r_result} show the comparison results of our method and existing methods on three datasets, CIFAR100, ImageNet100 and ImageNet-R. Our method outperforms various competing methods in most settings by a significant margin.
On the ImageNet100 dataset, our method achieves a final accuracy that is at least 1.58\% higher than other methods. Our method outperforms other methods by at least 1.08\% in final accuracy in the base0 setting experiment on the ImageNet-R dataset.
This shows that we have good effects in reducing forgetting and maintaining model stability by using textual information and parameter fusion. Compared with zero-shot Continual-CLIP, our method only adds a learnable linear layer to the model structure, but it also has a significant improvement. This shows that our method does not rely solely on the generalization of the pre-trained model to achieve good performance. \Cref{fig:curve} illustrates the decreasing trend of average accuracy as the number of classes increases for different datasets and settings. Our method can significantly mitigate forgetting.

The three prompt-based methods suffer from a severe imbalance of prompt usage in the base50 and base100 settings because they use the same number of prompts for each task by default. This leads to predictions of the model being biased towards the base classes, making it hard to inject new knowledge.
L2P and DualPrompt have surpassed the performance of the original model pre-trained on ImageNet21k on the imageNet-R dataset. However, CODA performs poorly on the three datasets, suggesting that its method may be highly coupled with a specific pre-trained model, such as ViT pre-trained on Imagenet21k.

SLCA achieves better performance on the initial task by fine-tuning the backbone and the classifier. 
Consequently, its experimental results in the base50 and base100 settings are relatively good.
However, the accumulated forgetting caused by fine-tuning the backbone becomes more obvious in long-sequence settings.

ADAM mainly trains an adapter model on the initial task, and then does not train the model on the subsequent tasks.
Therefore, its performance depends on the proportion of data in the initial task. Since the model is not trained on the subsequent tasks, it maintains the model stability but does not learn new knowledge. The low plasticity of the model affects the performance.
Due to the large image style difference in ImageNet-R, the dataset requires the model to learn more knowledge to adapt to the rich image style, and this method performs worse on this dataset than on the unified style cifar100 and imagenet100.

PROOF, designed for CLIP, expands the adapter with each task and uses cross-modal attention to fuse information from text and images.
However, it does not consider the effect of neighboring categories in CLIP classification. Even without exemplar and extended parameters, our approach still has advantages.

\begin{table*}[t]
    \caption{{Results on CUB200 (B0 Inc20) and ImageNet1K (B0 Inc100).}}
    \centering
    \label{tab:cub_and_img1k}
\resizebox{0.99\textwidth}{!}{
    \begin{tabular}{cccccccccc}
    \toprule%
    &  &PROOF &L2P ++ &DualPrompt    &CODA  &Continual-CLIP    &SLCA  &ADAM-Adapter   &ours \\
    \midrule
    \multirow{2}{*}{CUB200} &Avg    &{\bf 83.11} &71.90 &71.74  &66.61  &60.60  &73.30  &78.80  &\underline{83.04}\\
    &Last   &\underline{75.53}  &62.99  &62.14  &50.88  &51.16  &60.39  &70.61  &{\bf 76.34}\\
    \midrule%
    \multirow{2}{*}{ImageNet1K}&Avg    &76.23 &79.30 &\underline{79.39}  &76.99  &72.96  &79.10  &76.60  &{\bf 81.73}\\
    &Last   &65.26  &69.60  &\underline{69.79}  &66.96  &64.44  &68.27  &68.74  &{\bf 72.58}\\
    \bottomrule%
    
    \end{tabular}
}
\end{table*}

We also evaluate our approach on the large dataset ImageNet1K and the fine-grained dataset CUB200. The results are shown in \cref{tab:cub_and_img1k}. The experimental results show that our method still has advantages in the large data set. Especially on the CUB200, a more difficult fine-grained dataset, the performance of our method is much improved compared to the zero-shot performance of the CLIP.

In summary, our method does not expand the model as the task increases, while maintaining the learning of new data and alleviating forgetting in the incremental tasks.
More results are provided in the supplementary material.

\subsection{Ablation Studies and Other Analysis}
\begin{table}[t]
\begin{minipage}[t]{0.51\linewidth}
    \centering
    \caption{{Module ablation results on B0 Inc10 ImageNet100. SG denotes sampling a few old category features from the Gaussian distribution. $\mathcal{L}_{hinge}$(random) is random selection of class pairs for $\mathcal{L}_{hinge}$. The meaning of PF w/o MD is to use only \cref{eq:mask} for parameter fusion and MD means matrix decomposition.}}
    \label{tab:module_ablation}
\resizebox{0.99\textwidth}{!}{
    \begin{tabular}{cc}
    \toprule
    {Ablation}&{Last Accuracy $\uparrow$}\\
    \midrule
    \makecell[l]{Adapter-finetune + SG (Baseline)}                &73.80\\
    \midrule
    \makecell[l]{Baseline + $\mathcal{L}_{hinge}$(random)}        &74.08\\
    \makecell[l]{Baseline + $\mathcal{L}_{hinge}$}                &76.04\\
    \makecell[l]{Baseline + $\mathcal{L}_{hinge}$ + PF w/o MD}    &79.28\\
    \midrule
    \makecell[l]{Baseline + $\mathcal{L}_{hinge}$ + PF w/ MD (Full)}     &80.23\\
    \bottomrule
    \end{tabular}
}
\end{minipage}
  \hfill
\begin{minipage}[t]{0.45\linewidth}
    \centering
    \caption{The prediction results of the model under different ablation experiment settings for 50 test images of ``kingsnake''. Three other snakes are the adjacent new categories that are selected by text feature similarity. More similar results are provided in the supplementary material.}
    \label{tab:loss_ablation}
\resizebox{0.88\textwidth}{!}{
    \begin{tabular}{lcc}
    \toprule
                                    &w/o $\mathcal{L}_{hinge}$   &w/ $\mathcal{L}_{hinge}$\\
    \midrule
         kingsnake                  &25     &35 \\
         night snake                &11     &8 \\
         worm snake                 &2      &1 \\
         eastern hog-nosed snake    &10     &5 \\
         others                     &2      &1 \\
    \midrule
         accuracy                   &0.5     &0.7\\
    \bottomrule
    \end{tabular}
}
\end{minipage}
\end{table}

\minisection {Module ablation.}
\Cref{tab:module_ablation} shows the results of different components in our method. Random means random selection of class pairs, rather than using the distance of the textual feature. In this case, the accuracy is very close to the baseline, while using the full module with textual features improves the accuracy by 2.24\% over the baseline. This indicates that the text-guided selection of the nearest class is effective.
The performance of the parameter fusion method without matrix decomposition and the full parameter fusion method with matrix decomposition shows that the parameter fusion method with fine-grained importance matrix is effective and matrix decomposition further enhances it.

\begin{wrapfigure}[18]{r}{0.48\textwidth}
  \centering
    \includegraphics[width=0.48\textwidth]{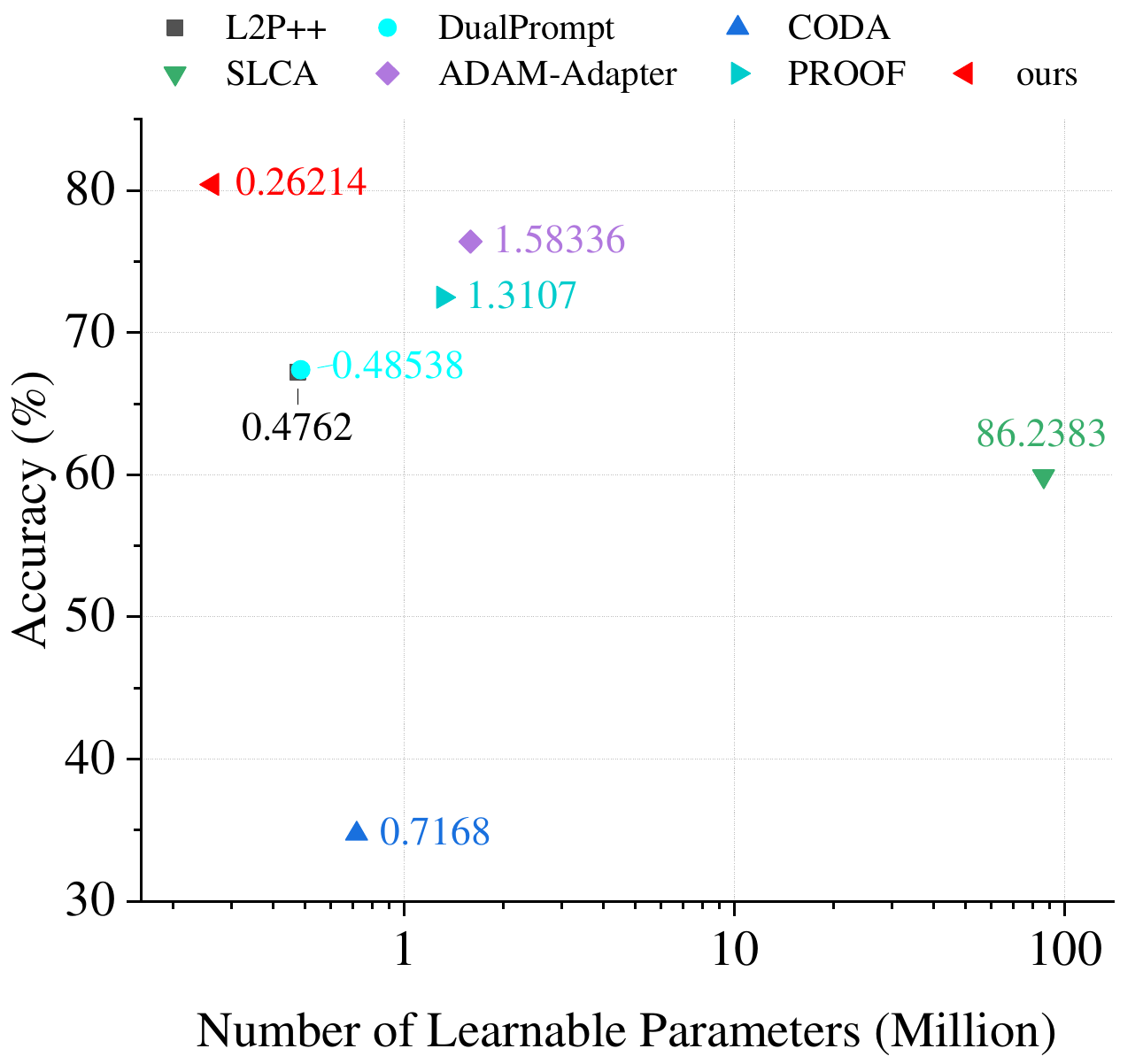}
  \caption{Comparison of methods in terms of accuracy and learnable parameters.}
\label{fig:parameter}
\end{wrapfigure}
\minisection {Training Cost Analysis.}
\Cref{fig:parameter} illustrates the comparison of our incremental parameter size with other methods on ImageNet100.
Using the same backbone, freezing the backbone parameters has much lower update costs than fine-tuning the entire model, as done by SLCA.
Among the methods that freeze the backbone and add learnable parameters, our method adds the fewest parameters.
Although we add a text encoder to the pure visual encoder method, because the label text is fixed, we only need to calculate the text features of the labels once during the entire training process and the number of labels is far lower than the data volume. 
Therefore, the cost of the text encoder is negligible compared to the visual encoder, which processes many images and multiple iterations.
For decomposed parameter fusion, we decompose the matrix only once after each task training. It is negligible compared to the training time.

\begin{figure}[t]
    \centering
    \begin{subfigure}[t]{0.305\linewidth}
        \includegraphics[width=\linewidth]{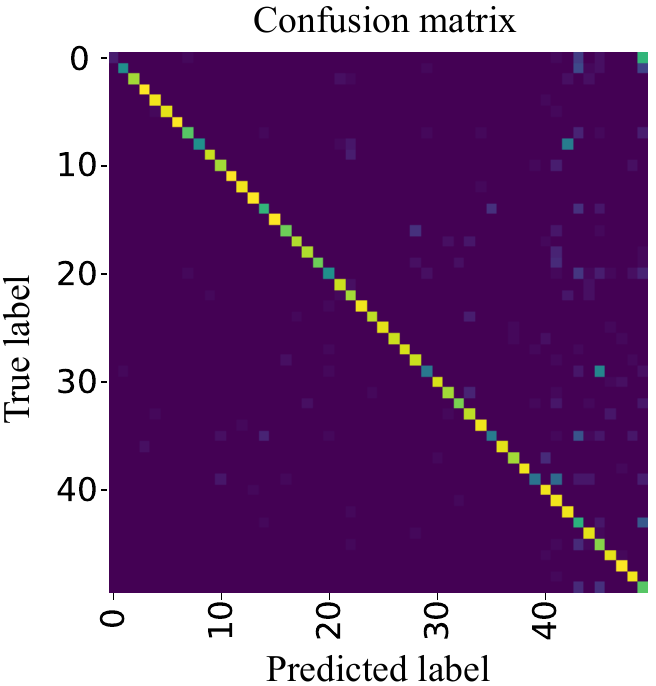}
        \caption{Baseline}
        \label{fig:cm_base}
    \end{subfigure}
    \begin{subfigure}[t]{0.305\linewidth}
        \includegraphics[width=\linewidth]{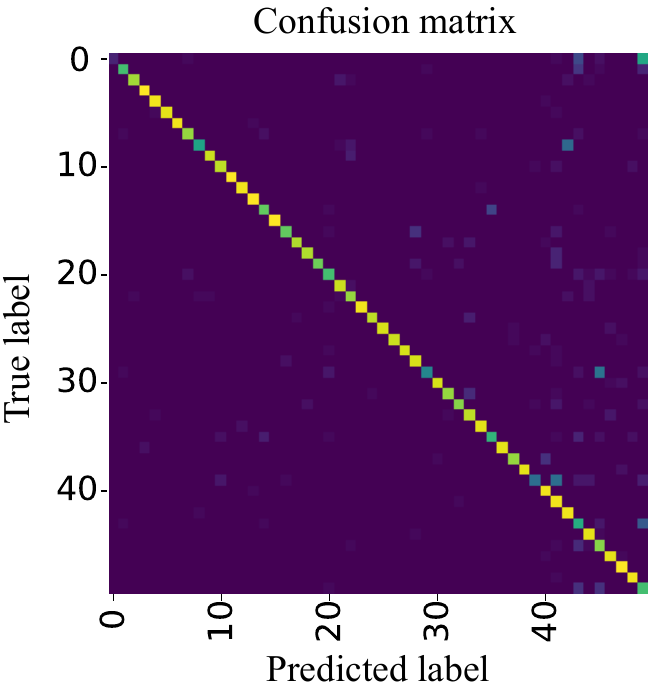}
        \caption{Baseline + $\mathcal{L}_{hinge}$}
        \label{fig:cm_our}
    \end{subfigure}
    \begin{subfigure}[t]{0.369\linewidth}
        \includegraphics[width=1\linewidth]{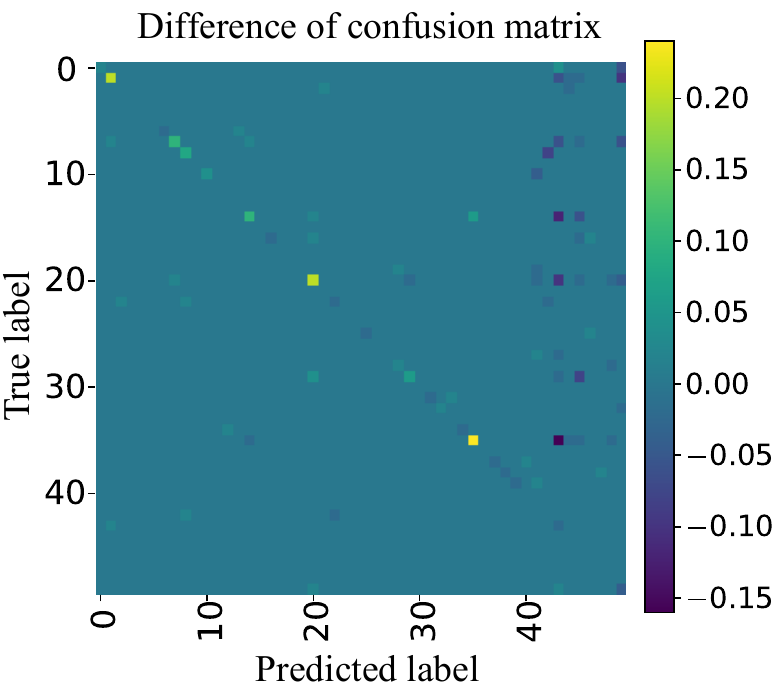}
        \caption{Difference between Fig. b and Fig. a (i. e. b - a)}
        \label{fig:diff_cm}
    \end{subfigure}
    \caption{The confusion matrixes of the first 5 tasks in the ImageNet100 B0 Inc10 experiment and their difference. We only show the first 5 tasks for better readability.}
    \label{fig:cm_all}
\end{figure}

\minisection {Neighboring categories classification.}
As shown in \cref{fig:cm_all}, our approach effectively corrects old categories misclassified into new categories.
\Cref{fig:diff_cm} shows that the number of samples misclassified to the new category decreases, the number of samples correctly classified increases, and the negative effect on other categories is small.
\Cref{tab:loss_ablation} shows some examples of the affected categories. It shows that the new categories that are close to the old categories selected by the similarity of the features of the text are the main cause of the wrong prediction of the old affected categories. The selected categories account for 23 of 25 erroneous predictions. See the supplementary material for more similar examples.
The $\mathcal{L}_{hinge}$ used for the selected categories can effectively reduce the erroneous prediction of the affected old categories.

\begin{wrapfigure}[10]{r}{0.48\linewidth}
\setlength{\abovecaptionskip}{0pt}
    \centering
    \includegraphics[width=0.99\linewidth]{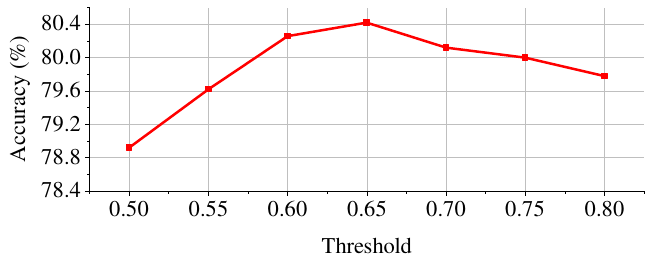}
    \caption{Experiment with different thresholds $\boldsymbol{\alpha}$ on the ImageNet100 B0 Inc10.}
    \label{fig:threshold}
\end{wrapfigure}

\minisection {Threshold ablation.}
In \cref{fig:threshold}, we study the effect of threshold $\boldsymbol{\alpha}$ for adjacent categories.
Since most distances are greater than 0.5, we gradually increase the threshold from 0.5.
As the threshold increases, more adjacent categories are selected, improving performance. However, a threshold that is too large will select many pairs of dissimilar categories, which will interfere with the classification and increase the computation. There is no benefit to choosing more different classes.
So we use 0.65 as the threshold for all experiments.

\minisection{Prompt-based methods with CLIP text encoder.}
For a fair comparison, we add the text encoder of CLIP to the classifier-independent prompt-based methods as a classifier. The results are shown in \cref{tab:clip_clf}.
The key and value of different tasks are relatively independent in L2P++ and DualPrompt, resulting in minimal change in the representation of the old classes. 
Using fixed CLIP text features instead of a trainable linear layer as the classifier may cause misclassification of the samples of old classes when the new class text feature is similar to the old class.
Hence, the performance of these two methods drops when using the CLIP text encoder.
CODA uses a weight to combine all prompt components enabling the old class to benefit from the newly expanded parameters. Performance is improved when using the information-rich text classifier. 
Compared with these methods, ours can make better use of the CLIP text encoder.

\minisection{Different ways to sample features for training.}
PASS~\cite{zhu2021pass} and Napa-vq~\cite{malepathirana2023napa_vq} are not designed based on pre-trained models and cannot be directly applied to pre-trained models.
Therefore, we replace the feature sampling and class separation process in our method with the PASS~\cite{zhu2021pass} and Napa-vq~\cite{malepathirana2023napa_vq} sampling methods as shown in \cref{tab:sample_methor}. PASS overlooks adjacent classes, while Napa-vq depends more on Na-vq modeling from scratch. Our strategy utilizes the CLIP model more effectively.

\begin{table}[t]
\begin{minipage}[t]{0.49\linewidth}
\centering
    \caption{Last accuracy under the B0 INC20 setting on ImageNet-R.}
\resizebox{\textwidth}{!}{
    \begin{tabular}{lcccc}
    \toprule
    & Backbone   &CODA&DualPrompt&L2P++\\
    \midrule
    & w/o text encoder &67.52  &75.77  &75.98  \\
    \midrule
    & w text encoder & 69.93 &72.30   &71.97  \\
    \bottomrule
    \end{tabular}
}
    \label{tab:clip_clf}


\end{minipage}
  \hfill
\begin{minipage}[t]{0.48\linewidth}
    \centering
    \caption{Results of different sampling under the B0 INC20 setting on CIFAR100.}
\resizebox{0.77\textwidth}{!}{
    \begin{tabular}{cccc}
    \toprule
    Method                                & \makebox[0.2\textwidth]{PASS}   & \makebox[0.2\textwidth]{Napa-vq} &  \makebox[0.2\textwidth]{ours}\\
    \midrule
         Avg                    &84.38             &84.23 &{\bf 86.19}\\
    \midrule
         Last                     &77.18      &77.22  &{\bf 79.04} \\
    \bottomrule
    \end{tabular}
}
    \label{tab:sample_methor}
\end{minipage}
\end{table}

\subsection{Comparison with Conventional Methods}

\begin{table}[t]
    \centering
    \caption{Comparison results of conventional methods on ImageNet-R B0 Inc20. The results of conventional methods are reproduced from PILOT~\cite{sun2023pilot} with the pre-trained weights on ImageNet21K.}
    \label{tab:conventional_label}
    \resizebox{0.6\textwidth}{!}{

    \begin{tabular}{cccccc}
    \toprule
    Method  &Coil~\cite{zhou2021coil}   &DER~\cite{yan2021der}  &iCaRL~\cite{rebuffi2017icarl}  &FOSTER~\cite{wang2022foster} &ours \\
    \midrule
    Avg     &80.48  &81.16  &72.76  &82.49  &{\bf 85.58}\\
    \midrule
    Last    &73.12  &75.10  &61.62  &76.00  &{\bf 82.28}  \\                          
    \bottomrule
    \end{tabular}
    }
\end{table}
Conventional methods, which did not use pre-trained models but with exemplars, also achieved good results in class incremental learning.
In \cref{tab:conventional_label}, we compare conventional methods that use the VIT-B/16 pre-trained on ImageNet21K for initialization. We have also evaluated these baselines with the same initialization as ours, but the performance is much worse. 
Our method still has a significant advantage over these state-of-the-art methods even without exemplar.

\section{Conclusion}
In this paper, we study the problem of incremental learning with pre-trained vision-language models. We find that introducing textual features of classes to adjust the representations of classes that are greatly affected by new data can effectively alleviate forgetting. Moreover, a simple linear adapter with a parameter fusion strategy can efficiently maintain model stability and reduce forgetting. Experiments demonstrate the effectiveness of our method. 
Manual threshold selection is a limitation.
Future work can design a mechanism to dynamically adapt the threshold and a mechanism to fuse the parameters more efficiently. 
The mutual influence of text and image can also be further explored.

\section*{Acknowledgements}
This work is funded by  
NSFC (NO. 62206135), Young Elite Scientists Sponsorship Program by CAST (NO. 2023QNRC001), Tianjin Natural Science Foundation (NO. 23JCQNJC01470), and the Fundamental Research Funds for the Central Universities (Nankai University). Computation is supported by the Supercomputing Center of Nankai University.

\bibliographystyle{splncs04}
\bibliography{main}

\clearpage

\appendix

\section{Classes of ImageNet100}
Following the setting of prior research~\cite{douillard2022dytox,thengane2022continualclip,yan2021der}, we select 100 categories from the ImageNet1k~\cite{deng2009imagenet} dataset as the ImageNet100 subset.
The class names of ImageNet100 are:

tench, goldfish, great white shark, tiger shark, hammerhead shark, electric ray, stingray, rooster, hen, ostrich, brambling, goldfinch, house finch, junco, indigo bunting, American robin, bulbul, jay, magpie, chickadee, American dipper, kite (bird of prey), bald eagle, vulture, great grey owl, fire salamander, smooth newt, newt, spotted salamander, axolotl, American bullfrog, tree frog, tailed frog, loggerhead sea turtle, leatherback sea turtle, mud turtle, terrapin, box turtle, banded gecko, green iguana, Carolina anole, desert grassland whiptail lizard, agama, frilled-necked lizard, alligator lizard, Gila monster, European green lizard, chameleon, Komodo dragon, Nile crocodile, American alligator, triceratops, worm snake, ring-necked snake, eastern hog-nosed snake, smooth green snake, kingsnake, garter snake, water snake, vine snake, night snake, boa constrictor, African rock python, Indian cobra, green mamba, sea snake, Saharan horned viper, eastern diamondback rattlesnake, sidewinder rattlesnake, trilobite, harvestman, scorpion, yellow garden spider, barn spider, European garden spider, southern black widow, tarantula, wolf spider, tick, centipede, black grouse, ptarmigan, ruffed grouse, prairie grouse, peafowl, quail, partridge, african grey parrot, macaw, sulphur-crested cockatoo, lorikeet, coucal, bee eater, hornbill, hummingbird, jacamar, toucan, duck, red-breasted merganser, goose.

\begin{table}[]
    \centering
    \caption{Comparison results of conventional methods on ImageNet-R B0 Inc20. The results of conventional methods are reproduced from PILOT~\cite{sun2023pilot} with the OpenAI CLIP pre-trained weights~\cite{radford2021clip}.}
    \begin{tabular}{cccc}
    \toprule
    {Method}
    &{Exemplar} 
    &Avg&Last\\
    \midrule
    Coil~\cite{zhou2021coil}        &\Checkmark     &45.55&20.65    \\
    DER~\cite{yan2021der}           &\Checkmark     &81.85&73.27    \\
    iCaRL~\cite{rebuffi2017icarl}   &\Checkmark     &77.25&64.52    \\
    FOSTER~\cite{wang2022foster}    &\Checkmark     &76.81&70.23     \\
    \midrule
    ours                            &\XSolidBrush     &85.58&80.28 \\
    \bottomrule
    \end{tabular}
    \label{tab:conventional_ini}
\end{table}

\begin{table}[]
    \centering
    \caption{The prediction results of the model under different ablation experiment settings for 50 test images of ``sea snake". ``Night snake", ``worm snake" and ``eastern hog-nosed snake" are the adjacent categories that are selected by calculating the text feature similarity among the newly added categories for the current task.}
    \begin{tabular}{lcc}
    \toprule
                                    &w/o $\mathcal{L}_{hinge}$   &$\mathcal{L}_{hinge}$\\
    \midrule
         sea snake                  &37     &42 \\
         night snake                &5     &2 \\
         worm snake                 &1      &0 \\
         eastern hog-nosed snake    &4     &1 \\
         others                     &3      &5 \\
    \midrule
         accuracy                   &0.74     &0.84\\
    \bottomrule
    \end{tabular}
    \label{tab:Neighboring_Categories_1}
\end{table}

\section{More Results of Experiments}
\subsection{Conventional Methods of CLIP Weight Initialization}
In \cref{tab:conventional_ini}, we compare the conventional methods that use the CLIP pre-trained VIT-B/16 model for initialization. Our method has a clear advantage.
\begin{figure*}[]
    \centering
    \includegraphics[width=\linewidth]{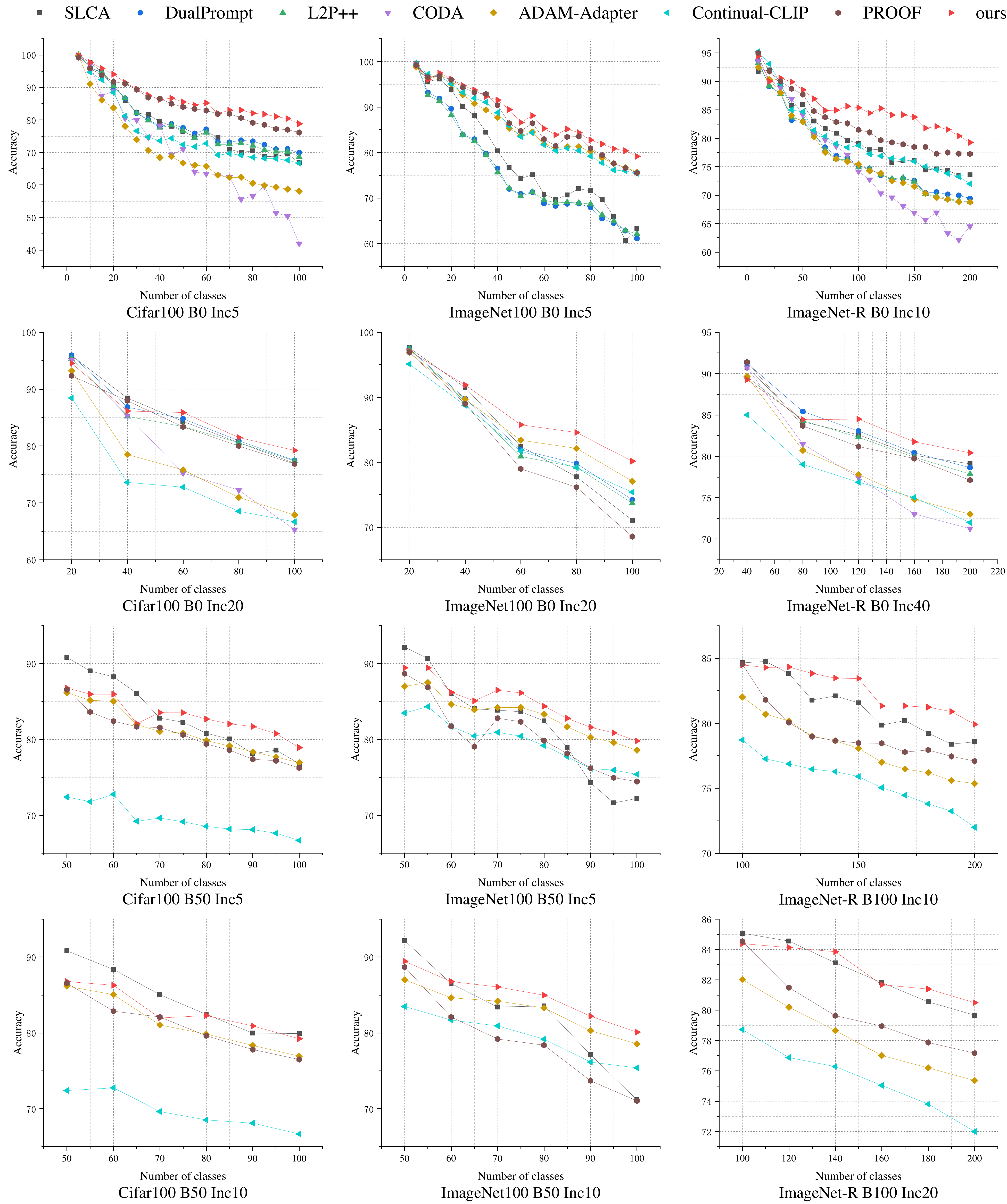}
    \caption{Accuracy curve of our method with other SOTA baselines on CIFAR100~\cite{krizhevsky2009cifar100}, ImageNet100~\cite{deng2009imagenet} and ImageNet-R~\cite{hendrycks2021imagenet-r}.}
    \label{fig:other_setting_curve}
\end{figure*}
\subsection{Accuracy Curve in More Settings}
As illustrated in \cref{fig:other_setting_curve}, our method outperforms the others in most scenarios. 
SLCA~\cite{zhang2023slca} fine-tunes the whole model. Because of that, SLCA performs better on the initial tasks in some settings, but its forgetting increases rapidly as the tasks increase.
\subsection{More Examples of Neighboring Categories}
Some experimental results of selected adjacent categories under different ablation settings are shown in \cref{tab:Neighboring_Categories_1}, \cref{tab:Neighboring_Categories_3} and \cref{tab:Neighboring_Categories_2}. Each category contains 50 test images. The results show that our method adjusts the representation of old categories with much less inference of the representation of new categories.
\begin{table}[]
    \centering
\centering
    \caption{Model prediction results of ``smooth newt" and ``newt" test images under different ablation experiment settings. ``Smooth newt" is the old category. ``Newt" is the adjacent category that is selected by calculating the text feature similarity among the newly added categories for the current task.}
    \begin{tabular}{lcc|cc}
    \toprule
                                    &\multicolumn{2}{c}{w/o $\mathcal{L}_{hinge}$}   &\multicolumn{2}{c}{$\mathcal{L}_{hinge}$}\\
                                        &smooth newt &newt &smooth newt&newt\\

    \midrule
         smooth newt                    &4      &0  &10 &2\\
         newt                           &22     &46 &16 &44\\
         others                         &24     &4  &24 &4\\
    \midrule
         accuracy                       &0.08   &0.92   &0.2    &0.88\\
    \bottomrule
    \end{tabular}
    \label{tab:Neighboring_Categories_3}
\end{table}

\begin{table}[]
    \centering
    \caption{Model prediction results of ``hen", ``quail" and ``goose" test images under different ablation experiment settings. ``Hen" is the old category. ``Quail" and ``goose" are the adjacent categories that are selected by calculating the text feature similarity among the newly added categories for the current task.}
    \begin{tabular}{lccc|ccc}
    \toprule
                                    &\multicolumn{3}{c}{w/o $\mathcal{L}_{hinge}$}   &\multicolumn{3}{c}{$\mathcal{L}_{hinge}$}\\
    & hen & quail &goose& hen & quail &goose\\
    \midrule
         hen                  &39     &0    &0  &43 &0  &0\\
         quail                &9      &50   &0  &5  &50 &0\\
         goose                &1      &0    &49 &1  &0  &48\\
         others               &1      &0    &1  &1  &0  &2\\
    \midrule
         accuracy             &0.78   &1    &0.98 &0.86 &1  &0.96\\
    \bottomrule
    \end{tabular}
    \label{tab:Neighboring_Categories_2}
\end{table}

\subsection{Comparison with RanPAC}
The performance comparison with RanPAC~\cite{mcdonnell2024ranpac} with OpenAI CLIP is in \cref{tab:Ranpac}. Settings denoted by * match the number of our learnable parameters (0.26M), compared to 1M-2M additional parameters of RanPAC.
While performance is close on CIFAR100, we excel on the more diverse ImageNet-R dataset.

\begin{table}[]
    \centering
        \caption{Comparison with RanPAC (10tasks).}
    \begin{tabular}{lcccc}
    \toprule
    &\multicolumn{2}{c}{CIFAR100} 
    &\multicolumn{2}{c}{ImageNet-R} \\
    &Avg &Last &Avg &Last \\
    \midrule
        RanPAC                    &87.15&80.96     &84.15 &77.66     \\
    \midrule
        RanPAC *                   &86.03&79.01     &80.22 &71.09 \\     
    \midrule
        ours                    &86.19 &79.04    &85.58 &80.28 \\
    \bottomrule
    \end{tabular}
    \label{tab:Ranpac}
\end{table}
As shown in \cref{tab:noise}, we also explored the effect of adding different proportions of random noise to the labels on ImageNet-R. It can be seen that our method is relatively robust to noise.

\begin{table}[]
    \centering
    \caption{Different proportions of noise in ImageNet-R (10tasks).}
    
    \begin{tabular}{lcccc}
    \toprule
                                    &0\%& 5\%   & 10\% &20\% \\
    \midrule
         RanPAC                    &77.66&77.54      &77.08  &70.7 \\
    \midrule
        ours                        &80.28&79.55  &79.25  &77.12 \\
    \bottomrule
    \end{tabular}
    \label{tab:noise}
\end{table}

\end{document}